Fida Ullah, Muhammad Ahmad, Muhammad Tayyab Zamir, Muhammad Arif, Grigori sidorov, Edgardo Manuel Felipe Riverón, and Alexander Gelbukh*

Centro de Investigación en Computación, Instituto Politécnico Nacional (CIC-PN), Mexico City 07738, Mexico
*Correspondence Autor: Alexander Gelbukh (gelbukh@cic.ipn.mx)

# EDU-NER-2025: Named Entity Recognition in Urdu Educational Texts using XLM-RoBERTa with X (formerly Twitter)

## Abstract

Named Entity Recognition (NER) plays a pivotal role in various Natural Language Processing (NLP) tasks by identifying and classifying named entities (NEs) from unstructured data into predefined categories such as person, organization, location, date, and time. While extensive research exists for high-resource languages and general domains, NER in Urdu—particularly within domain-specific contexts like education—remains significantly underexplored. This is Due to lack of annotated datasets for educational content which limits the ability of existing models to accurately identify entities such as academic roles, course names, and institutional terms, underscoring the urgent need for targeted resources in this domain. To the best of our knowledge, no dataset exists in the domain of the Urdu language for this purpose. To achieve this objective this study makes three key contributions. Firstly, we created a manually annotated dataset in the education domain, named EDU-NER-2025, which contains 13 unique most important entities related to education domain. Second, we describe our annotation process and guidelines in detail and discuss the challenges of labelling EDU-NER-2025 dataset. Third, we addressed and analyzed key linguistic challenges, such as morphological complexity and ambiguity, which are prevalent in formal Urdu texts. Fourth we employed 10 different experiment using 5-fold cross-validation by utilizing the power of machine learning including (SVM, LR, RF) using feature vector, two deep learning architectures (CNN, BiLSTM) using pre-trained FastText and GloVe embeddings, and three transformer-based models (BERT, RoBERTa, XLM-R) leveraging contextual embeddings to determine the best-fit model for the NER task and to capture dynamic meanings and contextual relationships within the textual data. Based on the analysis of the results the proposed model XLR-RoBERTa achieved remarkable cross-validation score of 98%, surpassing traditional methods (RF=0.89) by a margin of 10.11%. This work not only fills a significant gap in Urdu NER research but also establishes a robust foundation for future exploration of educational texts in under-resourced languages.

## Introduction

Named Entity Recognition (NER) is a process used to extract and categorize named entities (NEs) from unstructured textual, image, or video data, focusing primarily on identifying nouns or noun phrases classified into predefined categories such as person, organization, location, date, and time [1]. It serves as a foundational component for several downstream Natural Language Processing (NLP) tasks, including question answering [], machine translation, knowledge graph construction, query auto-completion systems, entity linking, and search engines [2–5], where the accurate extraction of NEs is crucial for precise classification. NLP also plays a significant role in analyzing the vast and diverse user-generated content on social media platforms like YouTube, Facebook, X, and Instagram, enabling the extraction of meaningful insights such as text classification [22], [23], [24], [25], [26], public sentiment, trending topics, and behavioral patterns. By facilitating tasks like sentiment analysis, topic modeling, and entity recognition, NLP empowers researchers, organizations, and policymakers to understand and respond to real-time discourse, bridging the gap between human language and machine understanding.

Urdu is Pakistan's primary language. It ranks as the world's twenty-first most widely spoken language [20] and is one of the most commonly used languages in the South Asian region. Pakistan's linguistic landscape encompasses more than 60 languages, including various regional dialects. Urdu, also known as the lingua franca, holds the status of the country's national language. While Urdu is readily understood by 75% of the population, it is the mother tongue of only 8%, according to statistical reports. The traditional and predominant use of the Nastaliq script in writing Urdu makes it one of the most challenging Arabic-script languages to learn. The Arabic script, distinguished by its unique directionality, is written from right to left across all languages that utilize this script system [6–9].

Urdu NER in the education domain faces two foundational challenges: the scarcity of annotated datasets and domain-specific linguistic complexities. No publicly available gold-standard corpus exists for educational texts (e.g., textbooks, academic reports), forcing researchers to rely on generic news-based datasets (e.g., Urdu NER Corpus by [1]) that poorly represent education-specific entities like course codes ("2024-میٹرک"), institutional roles ("پرنسپل ڈاکٹر حمید"), or hybrid terms ("وائس چانسلر آفس"). Domain adaptation is further hampered by Urdu’s morphological ambiguity in educational contexts—e.g., "لیب" could denote a laboratory (location) or a lab session (temporal event), while honorific-laden phrases ("پروفیسر صاحب") require specialized gazetteers. Code-mixing with English ("BS-IT پروگرام") and inconsistent transliterations ("فیصل vs فیصلہ") exacerbate tokenization errors. Without targeted annotation guidelines for educational entities (distinguishing "کیمسٹری ڈیپارٹمنٹ" as an organization vs. "کیمسٹری" as a subject), even state-of-the-art models fail to achieve robust F1-scores. Prioritizing crowd-sourced annotations of educational texts and rule-based normalization of academic jargon is critical to address this gap.

Existing studies on Urdu NER primarily employs three approaches: (1) rule-based, (2) machine learning and deep learning, and (3) hybrid methods. In the rule-based approach, text is matched using predefined grammatical rules or custom-built gazetteers. However, a key limitation of this method is its lack of portability across different domains, as the rules are highly specific to each domain. Additionally, creating rules for multiple domains demands significant manual effort, as well as domain-specific knowledge and language expertise, which can be time-consuming and labor-intensive [21]. We are particularly interested in the second approach, which involves machine learning and deep learning techniques, as it offers greater flexibility and scalability compared to rule-based methods, enabling more efficient handling of diverse domains without the need for manual rule development.

To achieve this objective , with the best of our knowledge we introduced a new and unique Urdu NER dataset named as Edu-UNER-2025 specifically focus on Education domain. For the construction of our EDU-NER-2025 dataset we sourced textual data from a popular social media platform such as Twitter. After the construction of the dataset we applied pre-processing techniques to purify the data and ready for machine learning process. After pre-processing we have employed foure different machine learning Support Vector machine (SVM), Logistic Regression (LR), and Random Forest (RF) using feature vector to extract the features from the textual data and we applied 2 popular deep learning method such as Convolutional Neural Network (CNN) and Bidirectional Long Short-Term Memory (BiLSTM) using advance pre-trained word embedding including FasText and Glove to extract the hidden pattern of NER dataset. Additionally we have employed 3 pre-trained language based transformer models including Bidirectional Encoder Representations from Transformers (BERT), Robustly Optimized BERT Pretraining Approach (RoBERTa) and XLM-RoBERTa using advance contextual embedding's.

This Study makes the following contributions:

- Existing studies on NER in Urdu primarily target general domains such as news articles and social media. To the best of our knowledge, there has been no prior research focused specifically on Named Entity Recognition (NER) within the education domain for the Urdu language.
- We built the detail annotation process and guidelines for Urdu our EDU-UNER-2025 and addresses and analyze key linguistic challenges such as morphological complexity and ambiguity, which are prevalent in formal Urdu texts.
- We have employed 5-folds of 10 different experiments using the combinations of traditional supervised learning using token-based feature extraction, deep learning using pre-trained FastText and GloVE word embeddings and transformer-based techniques using advanced contextual embeddings to find the best fit model for NER task.
- Our proposed model based on fine tuning (XLM-R) achieved 98% cross validation accuracy, representing a 10.11% performance improvement over traditional machine learning approaches (RF = 0.89%).

## Literature Review

Named entities encompass a range of elements such as people's names, organizations, locations, and events, and their identification is crucial for various NLP tasks [10]. NER across multiple languages, exploring the ingenious application of various models and techniques not only showcases the versatility of NER but also underline the importance of tailoring NER solutions to the idiosyncrasies of each language, ultimately enriching our capacity to decipher and extract valuable knowledge from a global tapestry of texts [11].
The scientific literature on NER reflects substantial research efforts dedicated to this domain, spanning multiple languages. Several notable NE-recognition approaches have emerged, including lexicon-based, rule-based, and machine-learning techniques [12].

Haq et al. [6] addressed the challenges in Urdu NER by utilizing deep learning based models to reduce the need for manual feature engineering, which is a limitation of existing systems. They employed CNNs to explore character-level features and joined them with word embeddings. For all this procedure they built a new Urdu dataset with five named entity classes and Evaluate their models on four benchmark datasets, and achieved a 6.26% increase in F1-score, thus advancing the field with more robust and automated feature learning.

Kanwal et al., [8] built the largest Urdu NER dataset with nearly 100k manually labeled entities, filling a major resource gap. They created six new Urdu word embeddings using fastText, Word2Vec, and GloVe. They were the first to apply deep learning (NN and RNN) for Urdu NER . Finally, they ran 32 experiments to explore how models, embeddings, and datasets affect performance, offering valuable insights and achieved the 72% accuracy.

In a scientific investigation, Malik et al. [13] delved into the development of a NER and Classification System (NERC) for the Urdu language using Artificial Neural Networks (ANN). Their study involved the creation of an Urdu Named Entity (NE) corpus known as the Kamran-PU-NE (KPU-NE) corpus. For the development of the Urdu NERC system, two distinct learning algorithms were employed:

the Hidden Markov Model (HMM) and the Artificial Neural Network (ANN). The outcomes of their experiments demonstrated notable differences in performance between the two approaches. Specifically, when using the Hidden Markov Model (HMM), the system achieved its highest precision, recall, and f-measure values at 55.98%, 83.11%, and 66.90%, respectively.

Mukund et al. [14] developed a bootstrapped model for Urdu text processing with four levels: POS tagging, NE tagging, word boundary segmentation, and post-processing. They used Conditional Random Fields (CRFs) for POS and NE tagging, improving accuracy with bootstrapped data. A bigram HMM handled word segmentation, refined by a probabilistic language model. Grammar rules and lexicon lookups were applied to enhance final tag accuracy.

Mukund et al. [15] developed a robust and customizable natural language processing (NLP) Model for Urdu language to support different tasks including topic detection and sentiment analysis. They combine different modules such as word segmentation, POS tagging, and name entities tagging, using bootstrapping and Hindi resource sharing to overcome data scarcity. The model also helps Urdu-to-English transliteration and has been integrated into a broader text-mining platform. Evaluation shows their system performs at or above current state-of-the-art levels.

Wahab et al [17] tackles the challenge of Named Entity Recognition (NER) in Urdu, a language with limited annotated data and structural complexities like lack of capitalization. The authors propose a CRF-based method using both language-dependent (POS tags) and language-independent (context windows) features. They also introduce a new manually annotated Urdu NER dataset (UNER-I). Experiments on both UNER-I and an existing dataset show that their method improves F1 scores by 1.5% to 3% over baseline models. The results highlight the value of their dataset in supervised learning for Urdu NER.

Biswas et al. [18] demonstrated The hybrid approach amalgamates rule-based and statistical methods, In their study, the authors skillfully fused the Maximum Entropy model with the Hidden Markov Model (HMM) approach, resulting in a substantial enhancement in the accuracy of recognizing Named Entities (NEs). However, to resolve some limitations in the Hybrid Oriya NER (NER) system, researchers should focus on to utilize various methods such as incorporating character-level features, utilizing domain adaptation techniques to achieve higher accuracy.

Gupta and Joshi et al. [19] Translation of text from one language to another is achieved through statistical machine translation, but there remain shortcomings due to insufficient language resources. Constructing a linguistic corpus entails the essential task of extracting multiword expressions (MWEs). To cover this gap, Gupta and Joshi (2022) [19], devised an innovative expression extraction method tailored for Urdu languages in the domain of natural language processing. They generated training corpora for language, employed CRF++ models, and achieved remarkable system accuracy rates of 96.62% for Urdu. Additionally, expanding the size of the training corpus in the future could enhance the efficiency and reliability of the system for automatically identifying and extracting multiword expressions (MWEs).

Unlike previous studies that focus on general-purpose or social media-based NER datasets for Urdu, our work introduces the first manually annotated Named Entity Recognition (NER) dataset in the education domain for the Urdu language. This dataset is uniquely designed to support domain-specific NLP applications and includes 13 distinct named entity classes, as illustrated in Figure 4. These classes were carefully defined to reflect the nuances and terminology used in educational texts. To the best of our knowledge, no prior research has developed an Urdu NER dataset tailored specifically to the education domain with such a rich and domain-aligned class structure.

**Methodology**

## Construction of dataset

This section will discuss the construction of dataset for our (Edu-NER-2025) dataset. Initially we develop a python code and utilized the Twitter API to retrieve nearly 30, 0000 Urdu tweets related to education, This study explores key terms such as تعلیم (education), کتاب (book), اساتذہ (teachers), طلبہ (students), and آن لائن (online), along with their related variations, to create a rich and meaningful EDU-NER-2025 dataset. The collected data spans a one-year period (2023-2024), ensuring its relevance and currency. To maintain the quality and accuracy of EDU-NER-2025, a team of domain experts was involved in defining and organizing an education-specific NER schema for Urdu. Our primary objective was to create a robust dataset related to Education field that will help to identify important named entities in social media discourse. Following the tweet-scraping process, we applied pre-processing techniques to eliminate irrelevant data and prepared corpus for manually annotation. After the corpus creation we each tweet was then labeled into one of 11 categories, such as person names, institute names, book titles, and locations as shown in Figure 4. Once our dataset was ready, we used it to train and fine-tune multiple models, including Machine Learning (ML), Deep Learning (DL), Transfer Learning (TL), and Large Language Models (LLM). Figure 1 shows the proposed methodology and design.

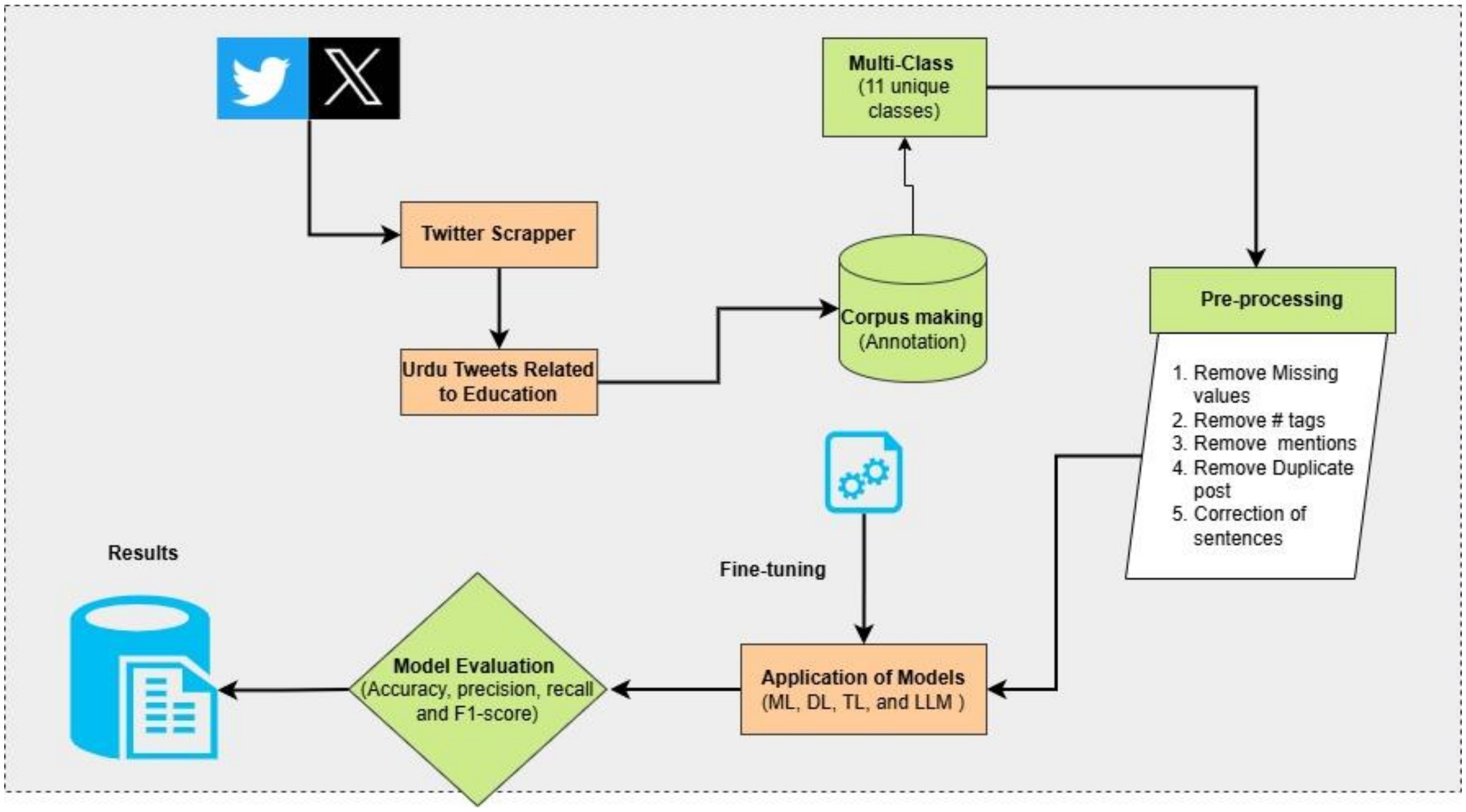

Figure 1. Proposed architecture and design.

## Annotations

Data annotation is the method of classifying samples to create high-quality training data for machine learning models. We perform data annotation procedure to accurately label to ensure that our dataset is suitable for training and evaluating NLP models. Due to the inherent noise and unstructured format of tweets, we applied manual annotation schema to improve accuracy by adding human-verified labels and reducing errors. This process allows us to structure the dataset effectively by identifying key entities such as person, location, organization, designation, book, event, course, date, number, time and other.

**Annotation guidelines**

Data annotation guideline is a set of instructions provided to annotators to help them to ensure consistent, accurate, and high-quality labeling of dataset. These guidelines help annotators understand what to label, how to label, and what to avoid when tagging entities in a dataset. In our task like NER (NER), annotation guidelines explain how to classify entities such as people, places, organizations and dates as shown in Table 1. We build following annotation guidelines to identify tweets that should be marked as NER detection.

**Be Consistent:** Always label the similar entity the same way throughout the dataset. Incorrect and inconsistent tagging is not allowed for example "University of Lahore" → ORGANIZATION always use this tag as organization.

**Mark the Full Entity:** read the full sentence and label the complete name of an entity. For example "Lahore University of Management Sciences" is labelled as→ ORGANIZATION.
**Handle Multiple Meanings Carefully:** read the full sentence and understand the meaning carefully in Urdu language some words have different meanings based on context. So choose the correct labels. For example "Apple launched a new iPhone" ایپل نے نیا آئی فون لانچ کیا → (ایپل) Apple = COMPANY "She ate an apple" → Apple = (Not an entity, do not label).

**Ignore Common Words & Generic Terms:** Only label specific names (people, places, organizations). For example """ پنجاب یونیورسٹی Punjab University" → is ORGANIZATION
**Handle Abbreviations & Different Spellings Properly:** Read the full sentence and convert the short word to full word for example If an entity has different spellings or abbreviations, label them the same way. For example یُو - او - ایل "UoL" یونیورسٹی آف لاہور (University of Lahore) → ORGANIZATION

**If You Are Unsure, Ask:** If annotators were not sure how to label an entity, do not guess. Instead, flag it for review so we can check it together.

**Final Label:** Double-check your annotations for accuracy. Always follow these rules to ensure high-quality data. If you have any doubts, ask before making a decision.

**Annotation procedure**

To ensure high-quality annotation, initially we carefully selected eight master's students as annotators. They were native speaker in Urdu language with sufficient knowledge of machine learning and NLP. First, we provided 200 samples for annotation. After carefully reviewing their labels, we found that five out of eight annotators annotate the nearly same labels. In the next round, we provided these five annotators another set of 200 posts for labeling. Upon evaluation, we observed that only three annotators consistently provided accurate and correct labels. To maintain data reliability, we selected these annotators for the final annotation of our NER dataset. This step-by-step selection helped us eliminate errors and inconsistencies. To continue this process, we individually created Google Forms to assess performance. If there was any misunderstanding during data annotation, we held a meeting to resolve the error.

**Inter Annotator Agreement**

We calculate inter-annotator agreement (IAA) to confirm the consistency and reliability of our NER dataset. Since three annotators labeled our dataset, it was crucial to calculate how much they agreed on the annotated labels. To measure the reliability of annotations, we calculated **Fleiss' Kappa**, which showed **79% agreement** among the annotators indicating substantial agreement as shown in Table 2. This high agreement score shows that our **annotation guidelines were clear** and that we chose **skilled annotators**.

**Tale 1. Fleiss' Kappa Interpretation.**

| Fleiss' Kappa Value (κ) | Level of Agreement |
|---|---|
| < 0.00 | Poor Agreement |
| 0.00 – 0.20 | Slight Agreement |
| 0.21 – 0.40 | Fair Agreement |
| 0.41 – 0.60 | Moderate Agreement |
| 0.61 – 0.80 | Substantial Agreement |
| 0.81 – 1.00 | Almost Perfect Agreement |

**Corpus characteristics and standardization**

The Figure 4 titled "Label Distribution of EDU-NER-2025" gives us a clear snapshot of how various named entity categories are distributed in our specialized educational dataset. It uses a logarithmic scale to show the wide range of frequencies more effectively. Among all labels, "Total Tokens" dominates with nearly 599,000 entries, followed by "Other" at around 508,000, indicating a vast amount of general or uncategorized text. Labels such as "Sentences" (18,454), "Organization" (22,176), and "Person" (15,188) appear with notable frequency, reflecting their common presence in educational content. Categories like "Course" (4,048), "Book" (3,294), "Date" (3,136), and "Time" (2,719) are relatively less frequent, while more specific labels like "Location" (11,131) and "Designation" (18,568) sit in the mid-range. The chart overall emphasizes the richness and variety of labeled data, suggesting the EDU-NER-2025 dataset is both comprehensive and well-suited for tasks requiring nuanced natural language understanding in an educational context.

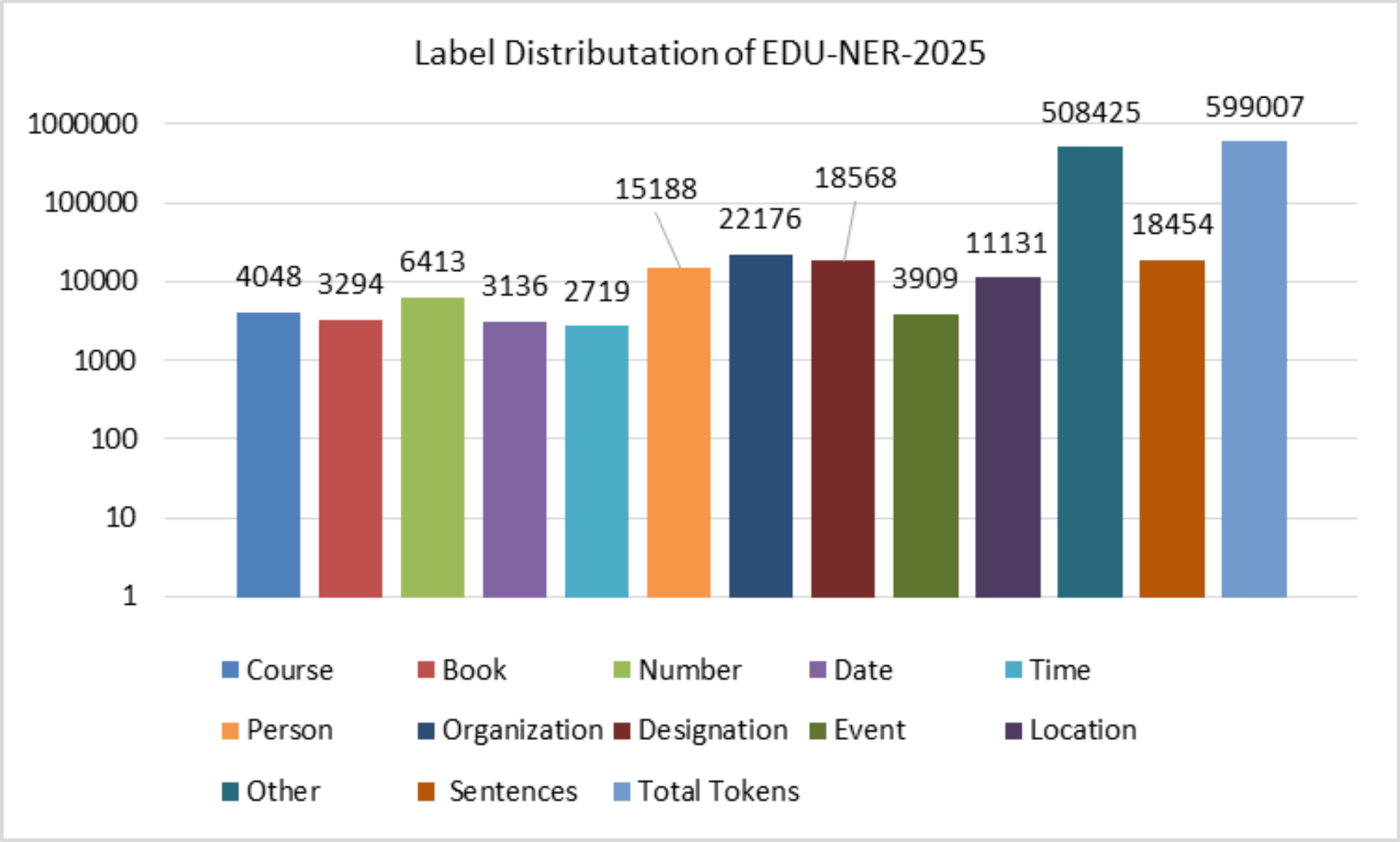

Figure 4. Label distribution of EDU-NER-2025 dataset.

The Figure 5 titled "Dataset Statistics" offers a comprehensive breakdown of key metrics for a Twitter-based dataset, using a logarithmic scale to accommodate the wide value range. The dataset includes 18,455 tweets, collectively containing 660,140 words, suggesting a substantial corpus for language analysis. On average, each tweet comprises 35.8 words, and roughly 166.3 characters, which aligns with the expanded tweet length limit that allows for more expressive content. The character count totals to a significant 3,061,987, highlighting the volume of text data available. Notably, the vocabulary size is 26,730 unique words, indicating a rich linguistic diversity that could support various natural language processing tasks such as sentiment analysis, topic modeling, or named entity recognition. This chart gives a solid sense of the dataset's scale and depth, making it a valuable resource for research or model training. While Figure 6 shows the word cloud image in the paper to visually represent the most frequently occurring terms in the dataset, providing an intuitive overview of the dominant themes and keywords.

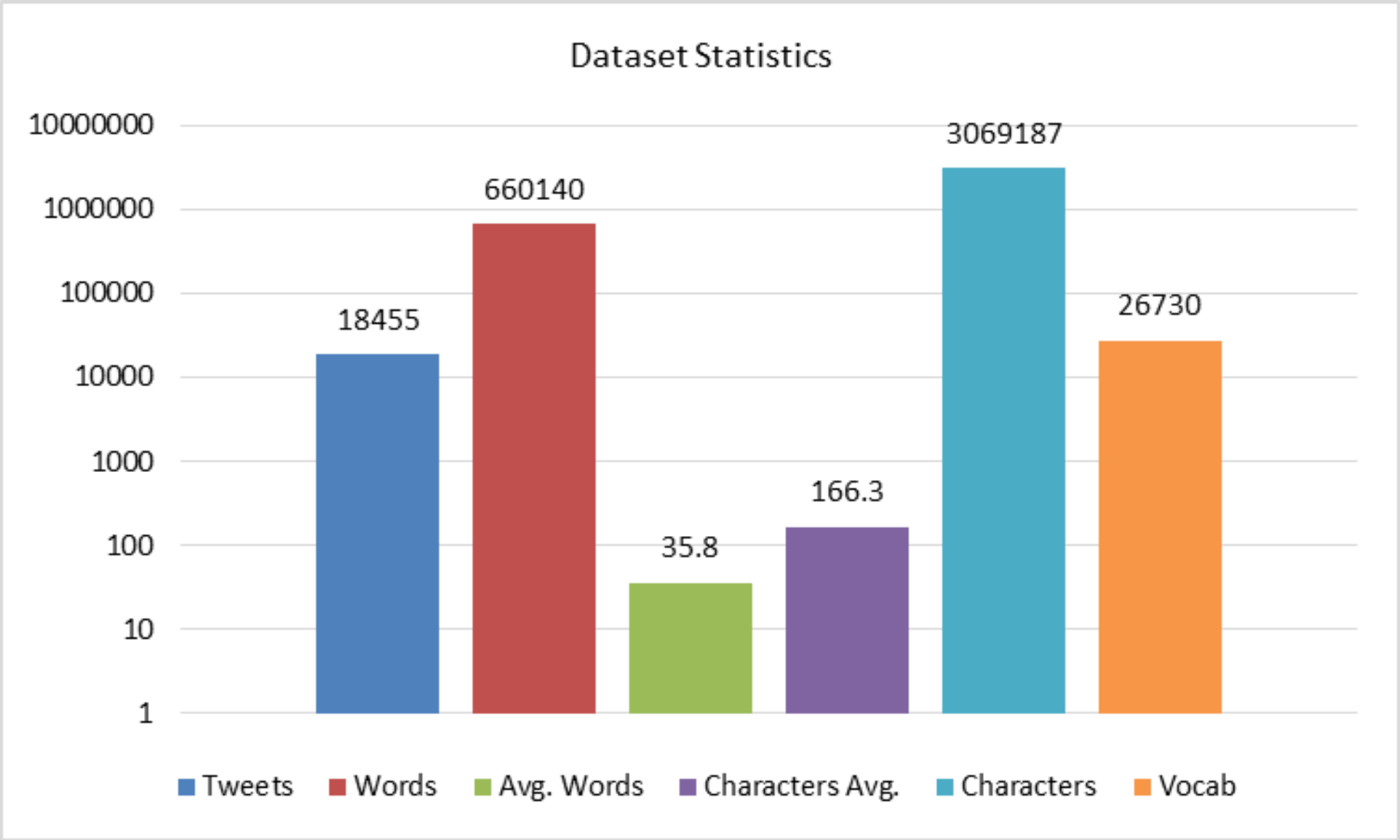

Figure 5. Dataset Statistics.

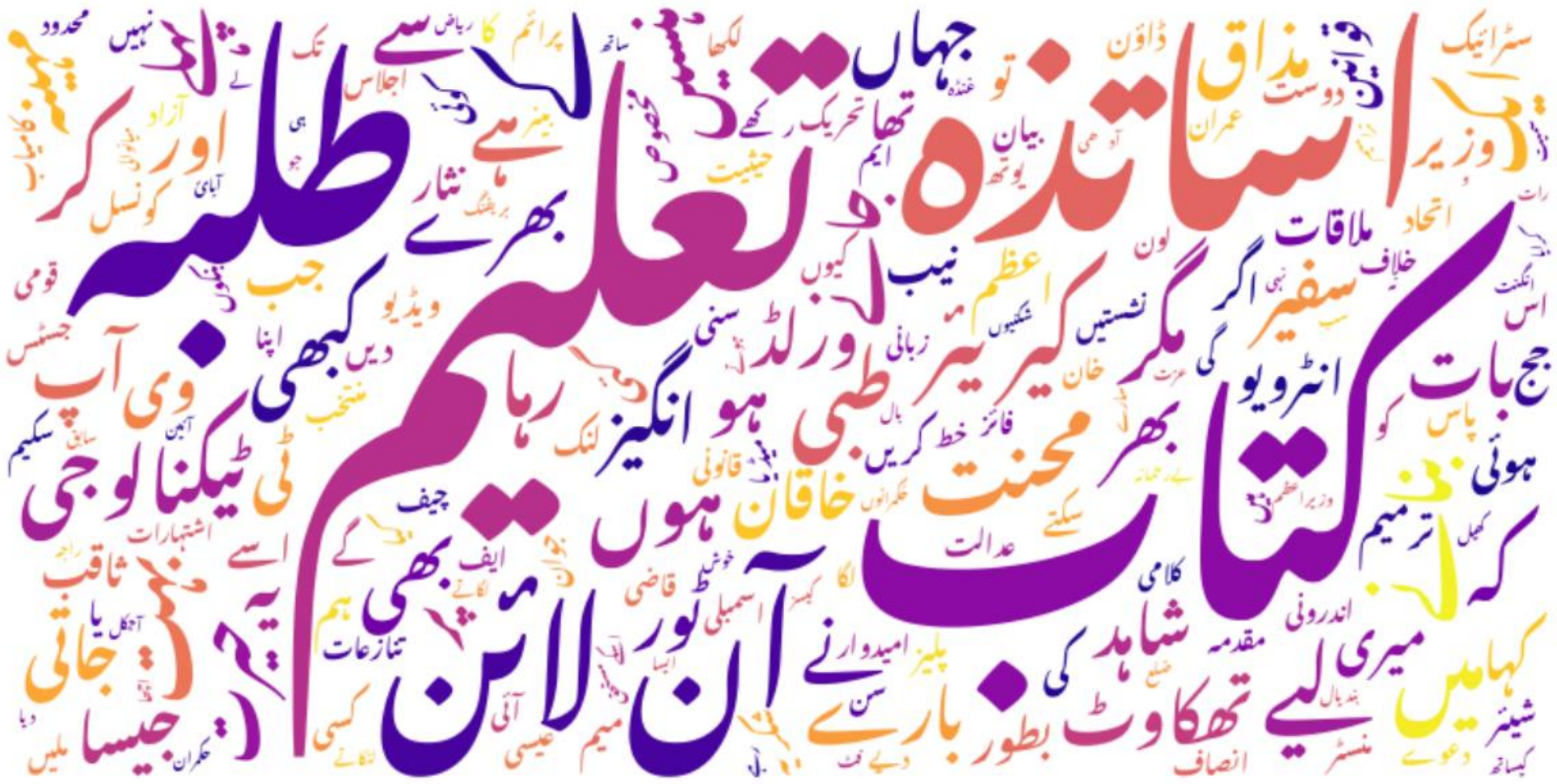

Figure 6. Word cloud showing the most frequently occurring words in the dataset.

**Pre-processing**

Data preprocessing is crucial step for an Urdu text dataset as social media data has multi language noise. For preparing and cleaning the data before applying NLP techniques like NER (NER). The process begins with the raw dataset, which contain special characters, punctuation marks, and short words that do not contribute meaningful information. These elements are removed to enhance the quality of the text. Next, the text is converted to lowercase to ensure uniformity and prevent duplication due to variations in capitalization. After that, the text undergoes tokenization, where it is divided into individual words or meaningful units. Following this, stop words (common words like "سے" "،میں" "،ہے") are removed to retain only essential words. Then, stemming is applied, which reduces words to their root forms (e.g., " → "پڑھائی پڑھ"") to help the model recognize different forms of the same word. After completing these steps, the final pre-processed dataset is clean, structured, and ready for training the NER model efficiently. Figure 2 shows the data pre-processing step used for our NER dataset.

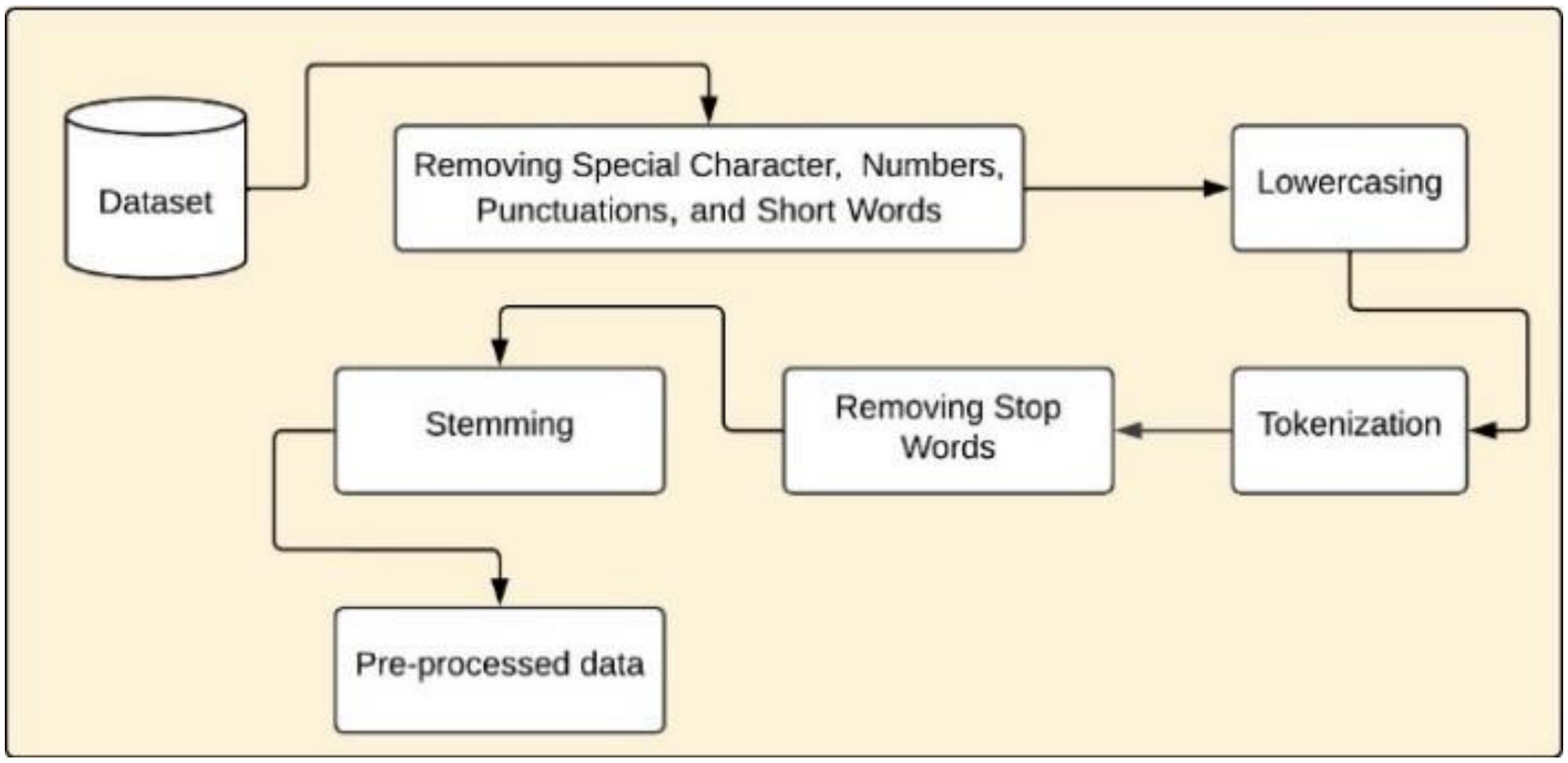

Figure 2. Data pre-processing approach

**Result and Discussion**

This section presents the outcomes of the experiments conducted using the methods described in the previous section. This section provides a detailed analysis of the performance metrics, highlights key observations, and interprets the results in light of the research objectives. The discussion aims to demonstrate how well the chosen approaches address the problem and what insights can be drawn from the obtained outcomes.

**Machine learning Results**

This section shows the results attained from the different machine learning models employed to our EDU-NER-2025 dataset. These models were evaluated based on the methodology outlined in the previous chapter. We utilized 4 different performance metric such as cross-validation score, precision, recall, and F1-score to assess the model's effectiveness.

The table 3 outlines the hyperparameters used for three machine learning models—Random Forest, Support Vector Machine (SVM), and Logistic Regression—each evaluated using 5-fold cross-validation (CV). For the Random Forest model, 100 decision trees (n_estimators) are used with no limit on tree depth (max_depth=None), and the criteria for splitting nodes and defining leaf nodes are set to the minimum values (min_samples_split=2, min_samples_leaf=1), while bootstrapping (bootstrap=True) is enabled to allow sampling with replacement. The SVM model uses a radial basis function ('rbf') kernel, with a regularization parameter C set to 1.0, gamma set to 'scale' which adapts to the input data, and probability=True to enable probability estimates. For Logistic Regression, the L2 regularization (penalty='l2') is applied with C=1.0 controlling the regularization strength, the 'liblinear' solver is chosen for optimization, and the maximum number of iterations (max_iter=1000) is set to ensure convergence. All models use 5-fold cross-validation to validate performance and reduce overfitting.

**Table 3. Fine-tuning parameters used for training Machine learning models using Feature Vector.**

| Model | Parameter | Value |
|---|---|---|
| **Random Forest** | n_estimators | 100 |
| | max_depth | None |
| | min_samples_split | 2 |
| | min_samples_leaf | 1 |
| | bootstrap | True |
| | **k-fold (CV)** | 5 |
| **SVM** | kernel | 'rbf' |
| | C | 1.0 |
| | gamma | 'scale' |
| | probability | True |
| | **k-fold (CV)** | 5 |
| **Logistic Regression** | penalty | 'l2' |
| | C | 1.0 |
| | solver | 'liblinear' |
| | max_iter | 1000 |
| | **k-fold (CV)** | 5 |

The figure 3 presents a comparative evaluation of three machine learning models—RF, SVM, and LR—on the NER task for the EDU-NER-2025 dataset, which focuses on Urdu text. Each model's performance is assessed using four key metrics: Precision (blue), Recall (red), F1-score (green), and C.V score (purple). RF and SVM both achieved the highest precision of 0.88 and recall of 0.89, with an F1-score of 0.86 and a consistent cross-validation score of 0.89, indicating stable performance. LR, while slightly lower in precision (0.87), maintained the same recall (0.89), F1-score (0.86), and C.V. score (0.89) as the other models. These results highlight the robustness and reliability of all three classifiers on the EDU-NER-2025 dataset, with RF and SVM showing a marginal edge in precision, suggesting they may better identify correct entities without sacrificing recall.

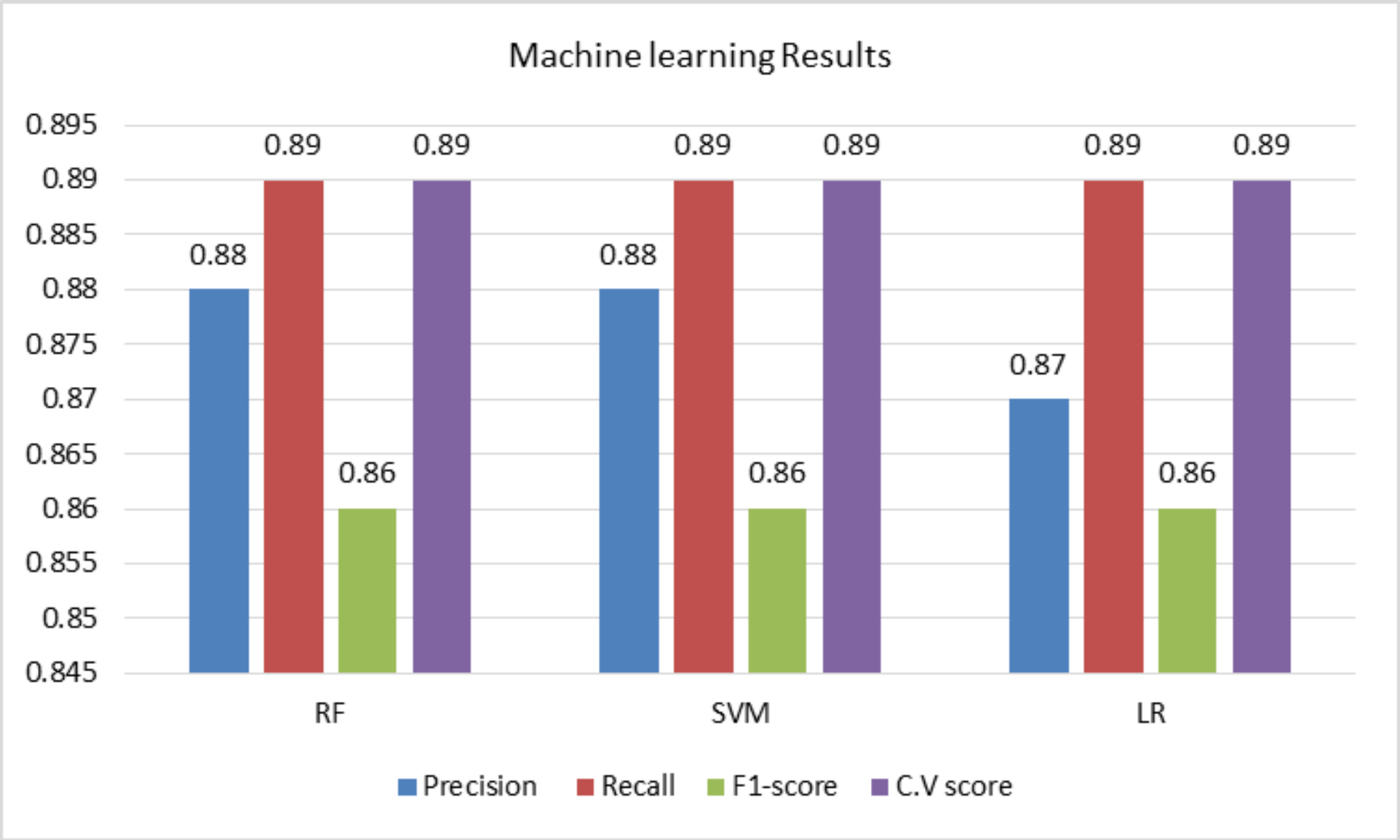

Figure 3. Results for the Machine learning models.

**Deep learning Results**

In this section, we explore the outcomes of two different deep learning models such as BiLSTM and CNN, combined with popular pre-trained word embedding's like fastText and GloVe. These models were trained on our dataset using cross validation score to see how well they could understand and classify the named entity based on the contextual information captured by the embeddings. To assess the model performance we utilized 4 different performance metric including cross-validation score, precision, recall, and F1-score.

The table 4 present the hyper-paramater tuning details of deep learning models utilized for the EDU-NER-2025 NER task. Both models were trained using pre-trained word embeddings—fastText and GloVe—to effectively capture semantic and syntactic relationships in the text. The training process for both models was carried out over 10 epochs with a batch size of 32 and a learning rate of 0.001, optimized using the Adam optimizer. We used categorical crossentropy as the loss function, suitable for multi-class classification tasks such as NER. To ensure the reliability and robustness of the models, we applied a 5-fold cross-validation strategy. A dropout rate of 0.5 was introduced to reduce overfitting. The BiLSTM model was configured with 128 hidden units, while the CNN model used 128 filters per layer. This consistent setup enabled a fair comparison of the two architectures in terms of their effectiveness on the NER task.

**Table 4. Fine-tuning parameters used for training BiLSTM and CNN models with FastText and GloVe embeddings**

| Parameter | BiLSTM | CNN |
|---|---|---|
| Embeddings Used | fastText, GloVe | fastText, GloVe |
| Number of Epochs | 10 | 10 |
| Batch Size | 32 | 32 |
| Learning Rate | 0.001 | 0.001 |
| Optimizer | Adam | Adam |
| Loss Function | Categorical Crossentropy | Categorical Crossentropy |
| Validation Strategy | 5-fold Cross Validation | 5-fold Cross Validation |
| Dropout Rate | 0.5 | 0.5 |
| Hidden Units/Filters | 128 (BiLSTM units) | 128 (filters per layer) |

The table 4 presents the results of the EDU-NER-2025 dataset using NER task, comparing the performance of two deep learning models—BiLSTM and CNN—using two different word embeddings: FastText and GloVe. When using FastText embeddings, both BiLSTM and CNN models achieved the same accuracy of 0.85, though BiLSTM slightly outperformed CNN in terms of F1-score (0.83 vs. 0.81), precision (0.85 vs. 0.83), and had equal recall (0.85). In contrast, when using GloVe embeddings, BiLSTM significantly outperformed CNN across all metrics, achieving a precision of 0.78, recall of 0.66, F1-score of 0.69, and C.V score of 0.66, while the CNN model performed poorly with a drastically low recall of 0.06, resulting in a very low F1-score of 0.11 and an overall C.V score of just 0.06. These results suggest that FastText embeddings are more effective for the EDU-NER-2025 dataset, and BiLSTM tends to be more robust across embedding types compared to CNN.

Table 4. Results for Deep learning models.

| Models | Precision | Recall | F1-score | C.V score |
|---|---|---|---|---|
| FastText | | | | |
| BiLSTM | 0.85 | 0.85 | 0.83 | 0.85 |
| CNN | 0.83 | 0.85 | 0.81 | 0.85 |
| GloVe | | | | |
| BiLSTM | 0.78 | 0.66 | 0.69 | 0.66 |
| CNN | 0.67 | 0.06 | 0.11 | 0.06 |

**Transfer learning Results**

This section highlights the results of three different advance language based transfer learning models using pre-trained contextual embeddings. The evaluation focuses on how effectively the model leveraged learned representations to perform the NER task. The discussion covers the overall performance, the model's generalization capability, and insights gained from using transfer learning in comparison to training models from scratch, though no direct baseline comparisons are made. We utilized multiple model to find the best solution for the NER task.

The table 5 outlines the hyperparameter settings used for fine-tuning three transformer-based models—bert-base-uncased, roberta-base, and xlm-roberta-base—for the Urdu NER task. All models leveraged transfer learning through their respective pretrained versions. A maximum sequence length of 128 tokens was set to efficiently handle input text. Training was conducted with a batch size of 16 and a learning rate of 2e-5, optimized using the AdamW optimizer, which is well-suited for transformer architectures. To prevent overfitting and improve generalization, a dropout rate of 0.1 and a weight decay of 0.01 were applied. Additionally, 500 warmup steps were used to gradually increase the learning rate at the beginning of training. Each model was trained for 5 epochs, and a 5-fold cross-validation strategy was employed to ensure robust and reliable performance evaluation across different data splits.

**Table 5. Fine-tuning parameters used for training in language based transformer models.**

| Models | Hyperparameter | Valuse |
|---|---|---|
| bert-base-uncased, roberta-base, xlm-roberta-base | Pretrained Model | Transfer leanring |
| | Max Sequence Length | 128 |
| | Batch Size | 16 |
| | Learning Rate | 2e-5 |
| | Optimizer | AdamW |
| | Weight Decay | 0.01 |
| | Dropout Rate | 0.1 |
| | Warmup Steps | 500 |
| | Number of Epochs | 5 |
| | Validation Strategy | 5-fold Cross Validation |

The Figure 3 presents the performance evaluation of three transformer-based models—BERT, RoBERTa, and XLM-RoBERTa—on our newly created Urdu NER dataset, named EDU-NER-2025. Each model was assessed using key performance metrics: precision, recall, F1-score, and cross-validation (C.V) score. The bert-base-uncased model achieved a consistent score of 0.97 across all metrics, demonstrating strong capability in identifying named entities in Urdu text. The roberta-base model followed closely with 0.96 across all metrics, showing slightly lower but still competitive performance. Notably, the xlm-roberta-base model outperformed the others with a score of 0.98 in all metrics, indicating its superior ability to handle multilingual data and effectively generalize across the complexities of the Urdu language. These results highlight the effectiveness of transformer-based models, particularly multilingual ones, in the Urdu NER task.

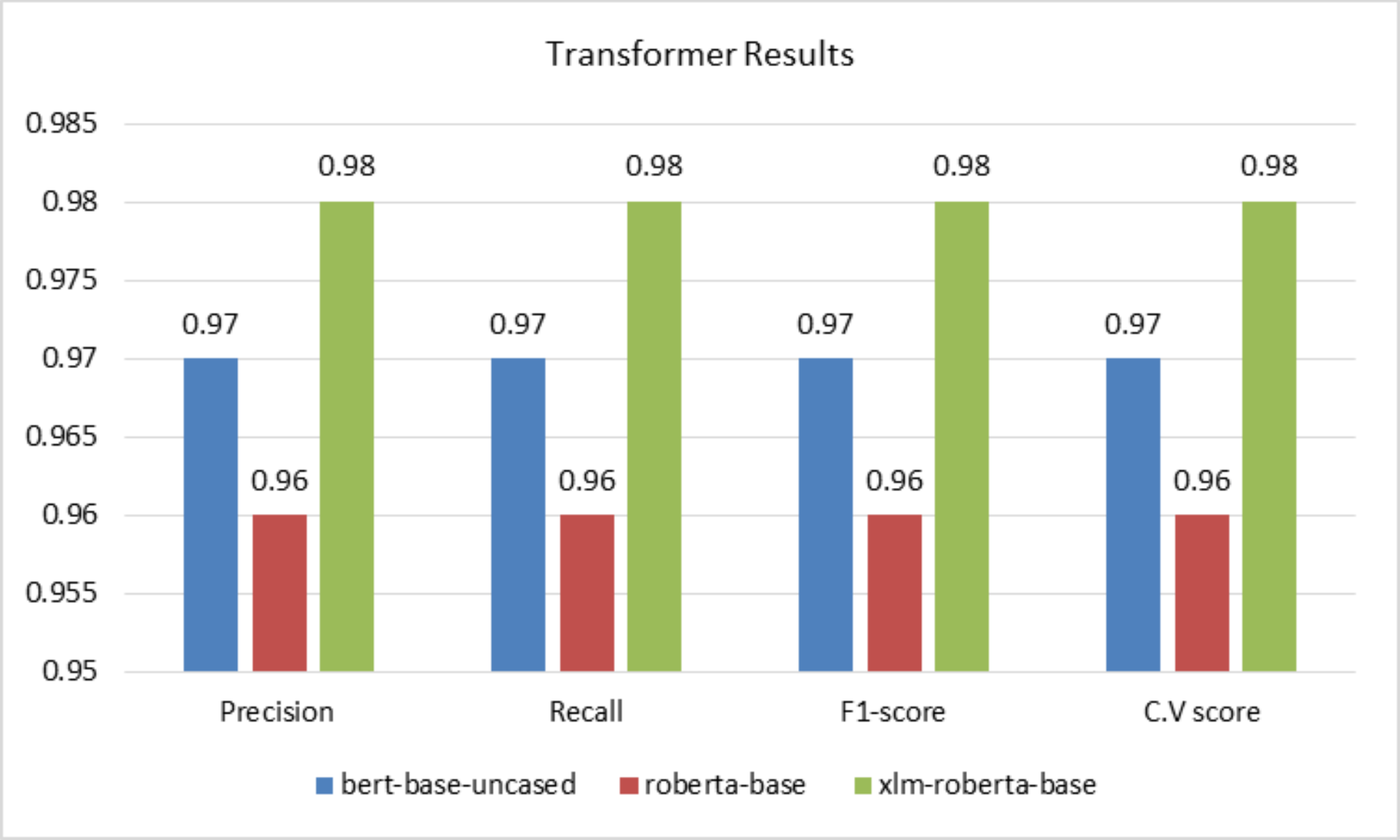

Figure 3. Results for Transfer learning models.

## Error Analysis

The Figure 4 highlights the top-performing models from three different learning approaches—machine learning, deep learning, and transfer learning—for the EDU-NER-2025 dataset using NER task. Among machine learning models, Random Forest (RF) stands out with strong metrics: a precision of 0.88, recall of 0.89, F1-score of 0.86, and an accuracy of 0.89. In the deep learning category, the BiLSTM model using FastText embeddings performs best, achieving 0.85 in both precision and recall, with an F1-score of 0.83 and accuracy of 0.85. However, the most superior results come from the transfer learning approach, where the XLM-RoBERTa-base model significantly outperforms all others, attaining an outstanding 0.98 across all evaluation metrics. This demonstrates that transfer learning, particularly with powerful multilingual transformers like XLM-RoBERTa, offers the most effective solution for the NER task in this setting.

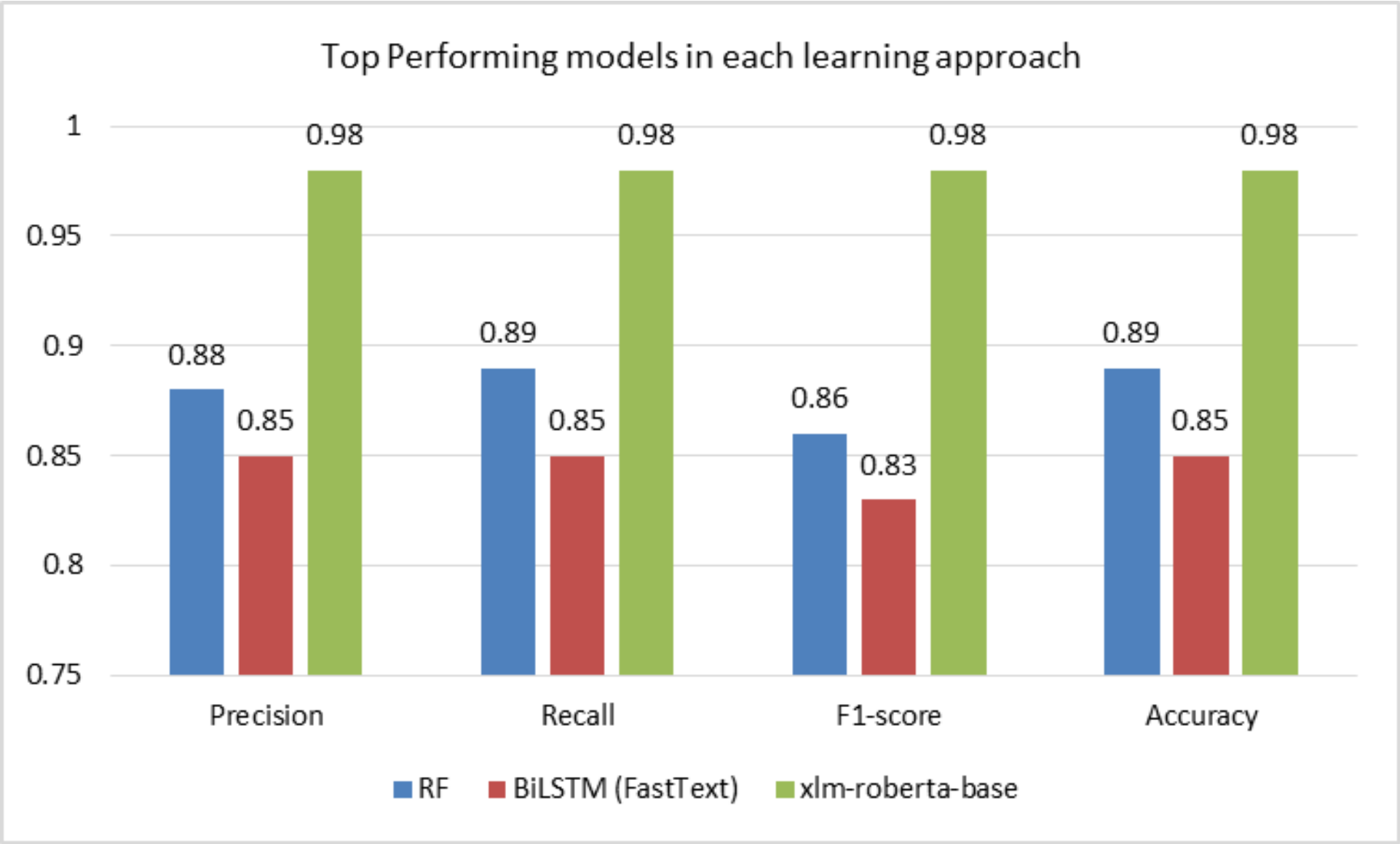

Figure 2. Top performing models in each learning approach.

The table 6 presents the class-wise evaluation metrics—precision, recall, F1-score, and support—for the Urdu NER task using the EDU-NER-2025 dataset. The model demonstrated consistently high performance across all entity types. Labels such as COURSE, EVENT, and OTHER achieved near-perfect scores with F1-scores of 0.99, indicating excellent precision and recall in identifying these entities. Categories like DESIGNATION, BOOK, and PERSON also showed strong results with F1-scores of 0.97, 0.97, and 0.95, respectively. While LOCATION and ORGANIZATION had slightly lower F1-scores of 0.92 and 0.94, they still reflect reliable entity recognition performance. The support column highlights the number of instances for each label, with OTHER being the most frequent (635,387 instances) and EVENT among the least (5,765 instances). Overall, the model maintained robust performance across a diverse set of entity categories, reflecting its effectiveness in handling complex NER tasks in the Urdu language.

Table 6. Class-wise Evaluation Metrics of the Proposed Model (XLM-RoBERTa).

| Label | Precision | Recall | F1-Score | Support |
|---|---|---|---|---|
| BOOK | 0.96 | 0.98 | 0.97 | 13,389 |
| COURSE | 0.98 | 0.99 | 0.99 | 16,792 |
| DATE | 0.96 | 0.95 | 0.95 | 6,162 |
| DESIGNATION | 0.97 | 0.97 | 0.97 | 31,351 |
| EVENT | 0.98 | 0.99 | 0.99 | 5,765 |
| LOCATION | 0.92 | 0.92 | 0.92 | 19,823 |
| NUMBER | 0.95 | 0.95 | 0.95 | 7,502 |
| ORGANIZATION | 0.94 | 0.95 | 0.94 | 46,689 |
| OTHER | 0.99 | 0.99 | 0.99 | 635,387 |
| PERSON | 0.95 | 0.95 | 0.95 | 40,646 |
| TIME | 0.95 | 0.96 | 0.95 | 6,083 |

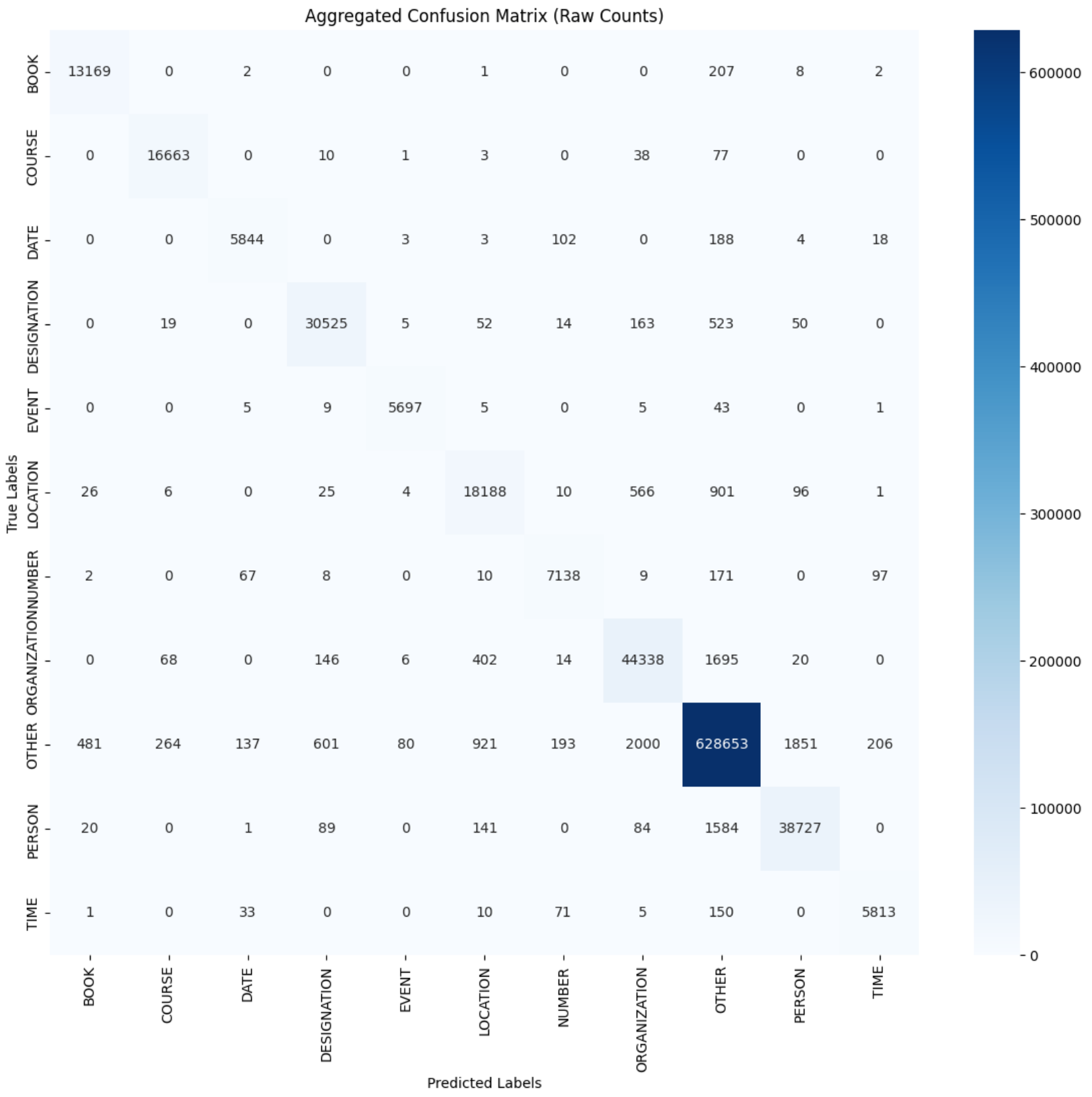

**Figure 3.** Confusion matrix of Top performing models (XLM-RoBERTa) utilized in this study.

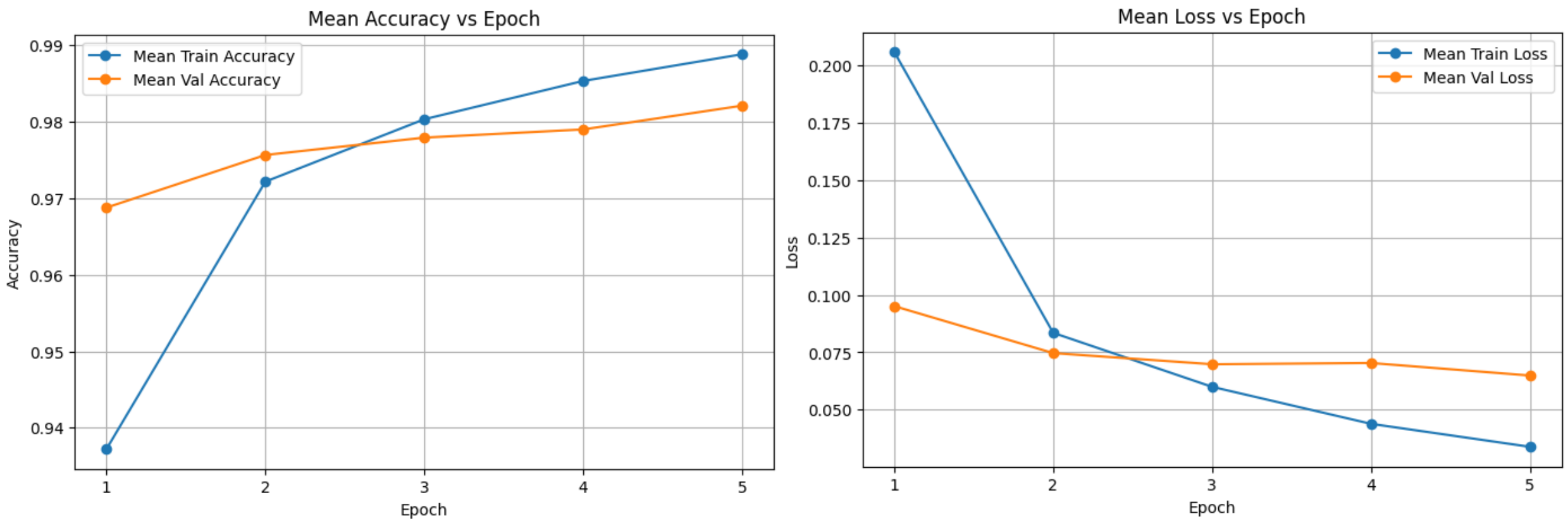

**Figure 4.** Training and validation performance of different epochs of proposed model XLM-R.

## Conclusion

This study presents *Edu-UNER-2025*, the first dedicated Named Entity Recognition (NER) dataset for the Urdu education domain, addressing a critical gap in low-resource language processing. By carefully annotating domain-specific entities from social media text and tackling unique linguistic challenges such as morphological ambiguity, code-mixing, and honorific expressions, we provide a robust resource to support downstream educational NLP applications. Extensive experimentation with machine learning, deep learning, and transformer-based models demonstrated the effectiveness of contextual embeddings, with fine-tuned XLM-RoBERTa achieving the highest performance of 98% accuracy—significantly outperforming traditional approaches.

In future work, we aim to expand *Edu-UNER-2025* to include additional educational genres such as academic reports, lecture transcripts, and textbooks to improve domain diversity and generalization. Additionally, incorporating Roman Urdu and transliterated content can enhance the dataset's adaptability to informal settings. We also plan to explore semi-supervised and active learning techniques for low-cost annotation and investigate cross-lingual transfer learning approaches to leverage high-resource educational corpora. Finally, integrating this NER system into downstream tasks like academic content classification, question answering, and knowledge graph construction could provide impactful real-world applications in Urdu language education technology.

**References**

1. Ehrmann, Maud, Ahmed Hamdi, Elvys Linhares Pontes, Matteo Romanello, and Antoine Doucet. "NER and classification in historical documents: A survey." *ACM Computing Surveys* 56, no. 2 (2023): 1-47.
2. Ullah, Fida, Alexander Gelbukh, Muhammad Tayyab Zamir, Edgardo Manuel Felipe Riverón, and Grigori Sidorov. "Enhancement of NER in Low-Resource Languages with Data Augmentation and BERT Models: A Case Study on Urdu." *Computers* 13, no. 10 (2024): 258.
3. Diefenbach D, Lopez V, Singh K, Maret P. Core techniques of question answering systems over knowl edgebases: a survey. Knowledge and information systems. 2018; 55(3):529–569. https://doi.org/10. 1007/s10115-017-1100-y
4. RogersA, GardnerM, Augenstein I. Qadatasetexplosion: A taxonomy of nlp resources for question answering and reading comprehension. ACMComputingSurveys.2023;55(10):1–45. https://doi.org/ 10.1145/3560260
5. Lewis P, OğuzB, RinottR, RiedelS, SchwenkH.MLQA: Evaluatingcross-lingual extractive question answering. arXiv preprint arXiv:191007475. 2019;.
6. Haq, R., Zhang, X., Khan, W., & Feng, Z. (2023). Urdu named entity recognition system using deep learning approaches. *The Computer Journal*, *66*(8), 1856-1869.
7. Khan, Wahab, Ali Daud, Fahd Alotaibi, Naif Aljohani, and Sachi Arafat. "Deep recurrent neural networks with word embeddings for Urdu NER." *ETRI Journal* 42, no. 1 (2020): 90-100.
8. Kanwal, S., Malik, K., Shahzad, K., Aslam, F., & Nawaz, Z. (2019). Urdu named entity recognition: Corpus generation and deep learning applications. *ACM Transactions on Asian and Low-Resource Language Information Processing (TALLIP)*, *19*(1), 1-13.
9. Ullah, Fida, Ihsan Ullah, and Olga Kolesnikova. "Urdu NER with attention bi-lstm-crf model." In *Mexican International Conference on Artificial Intelligence*, pp. 3-17. Cham: Springer Nature Switzerland, 2022.
10. Santana, B., Campos, R., Amorim, E., Jorge, A., Silvano, P., & Nunes, S. (2023). A survey on narrative extraction from textual data. Artificial Intelligence Review, 1-43.
11. Mulcaire, P., Kasai, J., & Smith, N. A. (2019). Polyglot contextual representations improve crosslingual transfer. arXiv preprint arXiv:1902.09697.
12. Sen, O., Fuad, M., Islam, M. N., Rabbi, J., Masud, M., Hasan, M. K., ... & Iftee, M. A. R. (2022). Bangla natural language processing: A comprehensive analysis of classical, machine learning, and deep learning-based methods. IEEE Access, 10, 38999-39044.

13. Malik, M. K. (2017). Urdu NER and classification system using artificial neural network. ACM Transactions on Asian and Low-Resource Language Information Processing (TALLIP), 17(1), 1-13.

14. Mukund, S., & Srihari, R. K. 2009. NE tagging for Urdu based on bootstrap POS learning. In Proceedings of the Third International Workshop on Cross Lingual Information Access: Addressing the Information Need of Multilingual Societies (CLIAWS3) (pp. 61-69).

15. Mukund, S., Srihari, R., & Peterson, E. (2010). An information-extraction system for Urdu---a resource-poor language. ACM Transactions on Asian Language Information Processing (TALIP), 9(4), 1-43.

16. Kotsiantis, S. B., Zaharakis, I. D., & Pintelas, P. E. (2006). Machine learning: a review of classification and combining techniques. Artificial Intelligence Review, 26, 159-190.

17. Khan, W., Daud, A., Shahzad, K., Amjad, T., Banjar, A., & Fasihuddin, H. (2022). NER Using Conditional Random Fields. Applied Sciences, 12(13), 6391

18. Biswas, S., Mohanty, S., Mishra, S. P.: A hybrid Oriya NER system: Integrating HMM with MaxEnt. In: 2009 Second International Conference on Emerging Trends in Engineering & Technology, pp. 639–643. IEEE (2009).

19. Gupta, V., & Joshi, N. (2022). Identification and extraction of multiword expressions from Hindi & Urdu language in natural language processing. International Journal of Advanced Technology and Engineering Exploration, 9(91), 807.

20. Fraser, P. D. J. (2009). *English: The prototypical world language for the twenty first century*. Lulu. com.

21. Singh, U., Goyal, V., & Lehal, G. S. (2012, December). Named entity recognition system for Urdu. In *Proceedings of COLING 2012* (pp. 2507-2518).

22. Ullah, F., Zamir, M. T., Ahmad, M., Sidorov, G., & Gelbukh, A. (2024). Hope: A multilingual approach to identifying positive communication in social media. In *Proceedings of the Iberian Languages Evaluation Forum (IberLEF 2024), co-located with the 40th Conference of the Spanish Society for Natural Language Processing (SEPLN 2024), CEUR-WS. org*.

23. Ahmad, M., Usman, S., Farid, H., Ameer, I., Muzammil, M., Hamza, A., ... & Batyrshin, I. (2024). Hope Speech Detection Using Social Media Discourse (Posi-Vox-2024): A Transfer Learning Approach. *Journal of Language and Education*, *10*(4 (40)), 31-43.

24. Usman, M., Ahmad, M., Ullah, F., Muzamil, M., Hamza, A., Jalal, M., & Gelbukh, A. (2025). Fine-Tuned RoBERTa Model for Bug Detection in Mobile Games: A Comprehensive Approach. *Computers*, *14*(4), 113.

25. Ahmad, M., Farid, H., Ameer, I., Muzamil, M., Jalal, A. H. M., Batyrshin, I., & Sidorov, G. (2025). Opioid Named Entity Recognition (ONER-2025) from Reddit. *arXiv preprint arXiv:2504.00027*.

26. Ahmad, M., Ameer, I., Sharif, W., Usman, S., Muzamil, M., Hamza, A., ... & Sidorov, G. (2025). Multilingual hope speech detection from tweets using transfer learning models. *Scientific Reports*, *15*(1), 9005.